\documentclass[11pt,letterpaper]{article}
\usepackage{emnlp2017}
\usepackage{times}
\usepackage{latexsym}
\usepackage{multirow}
\usepackage{pgfplots}
\usepackage{pgfplotstable}
\usepackage{graphicx}
\pgfplotsset{width=7cm, compat=1.4}
\usepackage{float}
\usepackage{filecontents}
\emnlpfinalcopy

\newcommand{\mysubsection}[1]{\vspace{0.3em} \noindent\textbf{#1}}

\title{Automatic Detection of Fake News}

\author{Ver\'{o}nica P\'{e}rez-Rosas$^1$, Bennett Kleinberg$^2$, Alexandra Lefevre$^1$  \\ \textbf{ Rada Mihalcea$^1$  } \\
  $^1$Computer Science and Engineering, University of Michigan \\ $^2$Department of Psychology, University of Amsterdam\\
  {\texttt {vrncapr@umich.edu,b.a.r.kleinberg@uva.nl,mihalcea@umich.edu}} }
\date{March 2017}

\begin{document}

\maketitle

\begin{abstract}
The proliferation of misleading information in everyday access media outlets such as social media feeds, news blogs, and online newspapers have made it challenging to identify trustworthy news sources, thus increasing the need for computational tools able to provide insights into the reliability of online content. In this paper, we focus on the automatic identification of fake content in online news. Our contribution is twofold. First, we introduce two novel datasets for the task of fake news detection, covering seven different news domains. We describe the collection, annotation, and validation process in detail and present several exploratory analysis on the identification of linguistic differences in fake and legitimate news content. Second, we conduct a set of learning experiments to build accurate fake news detectors. In addition, we provide comparative analyses of the automatic and manual identification of fake news.

\end{abstract}
\section{Introduction}

Fake news detection has recently attracted a growing interest from the general public and researchers as the circulation of missinformation online increases, particularly in media outlets such as social media feeds, news blogs, and online newspapers.  For instance, a recent report by the Jumpshot Tech Blog\footnote{https://www.jumpshot.com/data-facebooks-fake-news-problem/} found that Facebook referrals accounted for 50\% of the total traffic to fake news sites and 20\% total traffic to reputable websites. Since the majority of U.S. adults --62\%-- gets news on social media~\cite{Gottfried16}, being able to identify fake content in online sources is a pressing need. 

To date, computational approaches for fake news detection have relied on satirical news sources such as ``The Onion" and fact-checking websites such as "politiFact" and "Snopes". However, the use of these sources poses several challenges and potential drawbacks. For instance, using satirical content as a source for fake content can bring underlying confounding factors into the analysis, such as humor and absurdity. This is particularly the case for satirical news from ``The Onion", which has been used in the past to explore other text properties such as humor~\cite{Mihalcea05} and irony ~\cite{Wallace15}. On the other hand, fact-checking websites are usually constrained to a particular domain of interest, such as politics, and require human expertise, thus making it difficult to obtain datasets that provide some degree of generalization over several domains. 

In this paper, we develop computational resources and models for the task of fake news detection. We present the construction of two novel datasets covering seven different domains. One of the datasets is collected using a combination of manual and crowdsourced annotation efforts, while the second is collected directly from the web. Using these datasets, we conduct several exploratory analyses to identify linguistic properties that are predominantly present in fake content, and we build fake news detectors relying on linguistic features that achieve accuracies of up to 78\%. To place our results in perspective, we also compare the accuracy of our fake news detection  models with an empirical human baseline accuracy.

\section{Related Work}
To date, there are three important lines of research into the automated classification of genuine and fake news items. First, on a conceptual level, a distinction has been made between 'three types of fake news'~\cite{Rubin15}: serious fabrications (i.e. news items about false and non-existing events or information such as celebrity gossip), hoaxes (i.e. providing false information via, for example, social media with the intention to be picked up by traditional news websites) and satire (i.e. humorous news items that mimic genuine news but contain irony and absurdity). Here, we focus on the first category, serious fabrication, in the two domains of general news (in six different categories), as well as on celebrity gossip.

Second, attempts to differentiate satire from real news yielded promising results \cite{Rubin16}. The authors built a corpus of satire news (from \textit{The Onion} and \textit{The Beaverton}) and real news (\textit{The Toronto Star} and \textit{The New York Times}) in four domains (civics, science, business, soft news), resulting in a total of 240 news articles. The best classification performances were achieved with feature sets representing absurdity, punctuation, and grammar (each with an F1 score of 0.87).

Third, recently, a stylometric (i.e. writing-style) approach has been proposed for the identification of fake and genuine news articles \cite{Potthast17}. The investigation used the Buzzfeed dataset\footnote{http://github.com/BuzzFeedNews/2016-10-facebook-fact-check} of mainstream and hyperpartisan news articles of which the veracity was manually annotated. Stylometric features were, among others, character and stop word n-grams, readability indices, as well as features such as external links and the average number of words per paragraph. As a comparison, a topic-based feature set of a non-domain specific bag-of-words approach was used. The dataset used by \cite{Potthast17} consisted of 1,627 news articles that were obtainable from the original Buzzfeed dataset, including 299 fake news articles.  Although the stylometric approach was promising for the classification of hyperpartisan versus mainstream articles (accuracy: 0.75, compared to 0.71 for the topic-based feature set), both approaches were not able to differentiate fake from real news (accuracy: 0.55 and 0.52 for stylometric and topic-based feature sets, respectively).

Also related to our research is work done on the automatic identification of deceptive content, which has explored domains such as forums, consumer reviews websites, online advertising,  online dating, and crowdfounding platforms~\cite{Warkentin10,Ott11,Zhang08,Toma10,Shafqat16}. Linguistic clues such as self references or positive and negative words have been used to profile true tellers from liars~\cite{Newman03}. Other work has focused on analyzing the number of words, sentences, self references, affect, spatial and temporal information associated with deceptive content~\cite{Qin05}. Expressivity, informality, diversity and non-immediacy have also been explored to identify deceitful behaviors~\cite{Shafqat16}.  

\section{Fake News Datasets}
\label{lab:dataset}
As highlighted earlier, the datasets used in previous work have either relied on satirical news (e.g., ``The Onion''), which also have confounds such as humor or irony; or used fact-checking websites (e.g., ``politiFact'' or ``Snopes''), which are typically focused on only one domain (generally politics). We thus decided to construct two new datasets of fake news that cover several news domains and specifically model the deceptive property of fake news without major confounds. One dataset is collected via crowdsourcing, and covers six news domains; the second dataset is obtained directly from the web, and covers celebrity fake news.

\mysubsection{Guidelines for a Fake News Corpus.} 
In building a fake news dataset, we adhered to the nine requirements of a fake news corpus proposed by \cite{Rubin16}. Specifically, the authors suggested that such a corpus should (1) include both fake and real news items, (2) contain text-only news items, (3) have a verifiable ground-truth, (4) be homogeneous in length and (5) writing style, (6) contain news from a predefined time frame, (7) be delivered in the same manner and for the same purpose (e.g. humor, breaking news) for fake and real cases, (8) be made publicly available, and (9) should take language and cultural differences into account. 
In our work, to the extent possible, we aimed to address all of the above guidelines.\footnote{We  did not explicitly account for cultural differences since the primary aim was to build a fake news dataset that met criterion 1 to 8.} As outlined in the following, the ground-truth remains challenging since we cannot verify with absolute certainty whether all the content of real news items is in fact true.

\subsection{Building a Crowdsourced Dataset}

\mysubsection{Collecting Legitimate News.}
We started by collecting a dataset of legitimate news belonging to six different domains (sports, business, entertainment, politics, technology, and education). The news were obtained from a variety of mainstream news websites (predominantly in the US) such as the ABCNews, CNN, USAToday, NewYorkTimes, FoxNews, Bloomberg, and CNET among others. 

To ensure the veracity of the news, we conducted manual fact-checking on the news content, which included verifying the news source and cross-referencing information among several sources. Using this approach, we collected 40 news in each of the six domains, for a total of 240 legitimate news. 

\begin{table*}[ht]
\centering
\small
\begin{tabular}{p{7.5cm}|p{7.5cm}}
\hline
\multicolumn{1}{c|}{\sc Legitimate} & \multicolumn{1}{c}{\sc Fake  }\\ \hline
\textbf{Nintendo Switch game console to launch in March for \$299} The Nintendo Switch video game console will sell for about \$260 in Japan, starting March 3, the same date as its global rollout in the U.S. and Europe. The Japanese company promises the device will be packed with fun features of all its past machines and more. Nintendo is promising a more immersive, interactive experience with the Switch, including online playing and using the remote controller in games that don't require players to be constantly staring at a display. Nintendo officials demonstrated features such as using the detachable remote controllers, called "Joy-Con," to play a gun-duel game. Motion sensors enable players to feel virtual water being poured into a virtual cup. & \textbf{New Nintendo Switch game console to launch in March for \$99} Nintendo plans a promotional roll out of it's new Nintendo switch game console.  For a limited time, the console will roll out for an introductory price of \$99.  Nintendo promises to pack the new console with fun features not present in past machines. The new console contains new features such as motion detectors and immerse and interactive gaming. The new introductory price will be available for two months to show the public the new advances in gaming.  However, initial quantities will be limited to 250,000 units available at the sales price.  So rush out and get yours today while the promotional offer is running.\\
\hline 
\end{tabular}
\caption{Sample legitimate and crowdsourced fake news in the Technology domain}
\label{tab:sampleD1}
\end{table*}

\mysubsection{Collecting Fake News using Crowdsourcing.}
To generate fake versions of the news in the  legitimate news dataset, we make use of crowdsourcing via Amazon Mechanical Turk, which has been successfully used in the past for collecting deception data on several domains, including opinion reviews~\cite{ott-EtAl:2011:ACL-HLT2011}, and controversial topics such as abortion and death penalty~\cite{perezrosas-mihalcea:2015:EMNLP}. 

However, collecting deceptive data via AMT poses additional challenges on the news domain. First, the reporting language used by journalists might differ from AMT workers language (e.g., journalistic vs. informal style). Second, journalistic articles are usually lengthier than consumer reviews and opinions, thus increasing the difficulty of the task for AMT workers as they would be required to read a full news article and create a fake version from it. 

To address the former, we asked the workers to the extent possible to emulate a journalistic style in their writing. This decision was motivated by the 5th point of the fake news corpus guidelines described in section \ref{lab:dataset}, which suggests to obtain news with homogeneous writing style.  
To address the latter, we opted to working with smaller information units. Our approach consists of manually selecting a news excerpt that briefly describes the news article.\footnote{In many cases, this corresponded to the first 2-3 paragraphs in the document.} Thus, from the legitimate news dataset collected earlier, we manually extracted 240 news excerpts. The final dataset consists of 33,378 words. Each news excerpt has on average 139 words and approximately 5 sentences. 

We set up an AMT task that asked workers to generate a fake version of the provided news. Each hit included the legitimate news headline and its corresponding body. We instructed workers to produce both a fake headline and a fake news body within the same topic and length as the original news. Workers were also requested to avoid unrealistic content and to keep the names mentioned in the news. The fake news were produced by unique authors, as we allowed only a single submission per worker. We restricted the submission to workers located in the US as they might be more familiar with news published in the US media. In addition, we restricted participation to workers who maintained an approval rate of at least 95\% to reduce potential spam contributions.  

It took approximately five days to collect 240 fake news. Each hit was manually checked for spam and to make sure workers followed the provided guidelines. In general, we received few spam responses and most of the workers followed instructions satisfactorily; the only exceptions were a few cases where they provided only the headline or included unrealistic content.

Interestingly, we observed that AMT workers succeeded in mimicking the reporting style from the original news, which may be partly explained by typical verbal mirroring behaviors with drive individuals to produce utterances that match the grammatical structure of sentences they have recently read \cite{Ireland10}. 
This partially addresses our initial concern of authors reporting style being a source of noise while analyzing news generated by journalists and AMT workers.

\begin{table*}[ht]
\centering
\small
\begin{tabular}{p{7.5cm}|p{7.5cm}}
\hline
\multicolumn{1}{c|}{\sc Legitimate} & \multicolumn{1}{c}{\sc Fake  }\\ \hline
\textbf{Kim And Kanye Silence Divorce Rumors With Family Photo.} Kanye took to Twitter on Tuesday to share a photo of his family, simply writing, ``Happy Holidays." In the picture, seemingly taken at Kris Jenner's annual Christmas Eve party, Kim and a newly blond Kanye pose with their children, North, 3, and Saint, 1. After Kanye’s hospitalization, reports that there was trouble in paradise with Kim started brewing. But E! News shut down the speculation with a family source denying the rumors and telling the site, ``It's been a very hard couple of months." Kim remains out of the spotlight while Kanye is reportedly seeking outpatient treatment. Though Kim has yet to make a real return to social media herself, she's been spotted on Kanye’s page, as well as Khloe Kardashian's and Kylie Jenner's Instagrams and Snapchats. Kim and Ye were also photographed on a dinner date last week for the first time in a while, so things are looking up. & \textbf{Kim Kardashian Reportedly Cheating With Marquette King as She Gears up for Divorce From Kanye West.} Kim Kardashian is ready to file for divorce from Kanye West — but has she REALLY been cheating on him with Oakland Raiders punter Marquette King? The NFL star seemingly took to Twitter to address rumors that they've been getting close amid Kanye's mental breakdown, which were originally started by sports blogger Terez Owens. While he doesn't appear to confirm or deny an affair, her reps said there is ``no truth whatsoever" to the reports and labeled the situation "fabricated." As In Touch previously reported, Kim has been speaking with famed divorce attorney Laura Wasser and asked for documents to be drawn up. It has yet to be confirmed if Laura, who is also a friend of the reality star, will represent Kim during the proceedings. An insider blames the rapper's paranoia as a reason for the demise of their marriage. "Kim is miserable and wants this marriage to be over," says the source. 
\\
\hline 
\end{tabular}
\caption{Sample legitimate and web fake news in the Celebrity domain}
\label{tab:sampleD2}
\end{table*}

The final set of fake news consists of 31,990 words. Each fake news has on average 132 words and approximately 5 sentences. Table \ref{tab:sampleD1} shows a sample fake news, along with its legitimate version, in the technology domain. 

Throughout the rest of the paper, we refer to this crowdsourced dataset as FakeNewsAMT. 

\subsection{Building a Web Dataset}

We collected a second dataset of fake news from web sources following similar guidelines as in the previous dataset. However, this time, we aimed to identify fake content that naturally occurs on the web. We opted for collecting news from public figures as they are frequently targeted by rumors, hoaxes, and fake reports. We focused mainly on celebrities (actors, singers, socialites, and politicians) and our sources include online magazines such as Entertainment Weekly, People Magazine, RadarOnline, among other tabloid and entertainment-oriented publications. The data were collected in pairs, with one article being legitimate and the other fake. In order to determine if a given celebrity news was legitimate or not, the claims made in the article were evaluated using gossip-checking sites such as "GossipCop.com", and were cross-referenced with information from other sources.

During the initial stages of the data collection, we noticed that celebrity news tend to center on sensational topics that sources believe readers want to read about, such as divorces, pregnancies, and fights. Consequently, celebrity news tends to follow certain celebrities more than others further leading to an inherent lack in topic diversity in celebrity news. To address this issue, we evaluated several sources to make sure we obtain a diversified pool of celebrities and topics. Upon beginning the data collection procedure using these guidelines, another characteristic surfaced: several pairs contained nearly the same information with similar lexicon and reporting style, with differences being as simple as just negating the false news. For example, the following headlines correspond to a news pair where the legitimate version only negates the fake version: ``Aniston gets into fight with husband" (fake) and ``Aniston did NOT get into fight with husband" (legitimate). To address this issue, we sought to identify related news that still followed the fake-legitimate pair property while being sufficiently diverse in lexicon and tone. In the former example, the fake news was paired with an article titled ``Aniston and Husband enjoy dinner" that was published on the date of the alleged fight.

Using this approach, we collected 100 fake news articles and 100 legitimate news articles in the celebrity domain. The final fake news set has an average of 399 words and 17 sentences per article, for a total of 39,940 words. The corresponding legitimate news set has an average of 709 words and 33 sentences per article, for a total of 70,975 words. \ref{tab:sampleD2} shows an example of an article pairing in the dataset. 

Throughout the rest of the paper, we refer to this web dataset as Celebrity.

\begin{table*}[ht!]
\small
\centering
\begin{tabular}{l|l|ccc|ccc}
                        \hline
                        \multirow{ 2}{*}{Features (number of features)}&    \multirow{ 2}{*}{Acc.}      & \multicolumn{3}{c|}{\sc Legitimate} & \multicolumn{3}{c}{\sc Fake}                                 \\ \cline{3-8}
                        &  & P      & R      & F1     & P       & R      & F1           \\ \hline
Punctuation (11)             & 0.71     & 0.73   & 0.66   & 0.69   & 0.69    & 0.76   & 0.72        \\
LIWC - Summary  (7)      & 0.61     & 0.63   & 0.54   & 0.58   & 0.60    & 0.68   & 0.64          \\
LIWC - Linguistic processes (21)    & 0.67     & 0.66    & 0.67   & 0.66   & 0.67    & 0.66   & 0.66      \\
LIWC - Psychological processes (40)    & 0.56     & 0.56   & 0.57   & 0.56   & 0.56    & 0.56   & 0.55          \\
Complete LIWC (79)           & 0.70     & 0.70   & 0.71   & 0.70   & 0.71     & 0.70   & 0.70       \\
Readability  (26)     & 0.78     & 0.82   & 0.72   & 0.77    & 0.75     & 0.84    & 0.79 \\
Ngrams (651)      & 0.62     & 0.63   & 0.62   & 0.62    & 0.62     & 0.63    & 0.62          \\
Syntax (1375)                    & 0.65     & 0.66   & 0.63   & 0.64    & 0.64     & 0.67    & 0.65 \\ \hline
All Features (2131)    &  0.74     & 0.75   & 0.73   & 0.74    & 0.74     & 0.75    & 0.74  \\
\hline
\end{tabular}
\caption{Classification results FakeNews dataset collected via crowdsourcing. }
\label{tab:results_amt}
\end{table*}

\section{Linguistic Features}

To build the fake news detection models, we start by extracting several sets of linguistic features:

\mysubsection{\textit {Ngrams.}} We extract unigrams and bigrams derived from the bag of words representation of each news article. To account for occasional differences in content length, these features are encoded as tf-idf values.

\mysubsection{\textit {Punctuation.}} Previous work on fake news detection \cite{Rubin16} as well as on opinion spam \cite{ott-EtAl:2011:ACL-HLT2011} suggests that the use of punctuation might be useful to differentiate deceptive from truthful texts. We construct a punctuation feature set consisting of eleven types of punctuation derived from the Linguistic Inquiry and Word Count software (LIWC, Version 1.3.1 2015) \cite{pennebaker2015development}. This includes punctuation characters such as periods, commas, dashes, question marks and exclamation marks. 

\mysubsection{\textit {Psycholinguistic features.}}  We use the LIWC lexicon to extract the proportions of words that fall into psycholinguistic categories. LIWC is based on large lexicons of word categories that represent psycholinguistic processes (e.g., positive emotions, perceptual processes), summary categories (e.g., words per sentence), as well as part-of-speech categories (e.g., articles, verbs). Previous work on verbal deception detection showed that LIWC is a valuable tool for the deception detection in various contexts (e.g., genuine and fake hotel reviews, \cite{ott-EtAl:2011:ACL-HLT2011,ott2013negative}; prisoners' lies \cite{bond2005language}). In our work, we cluster the single LIWC categories into the following feature sets: summary categories (e.g., analytical thinking, emotional tone), linguistic processes (e.g., function words, pronouns), and psychological processes (e.g., affective processes, social processes).

We also test a combined feature set of all the LIWC categories (including punctuation).\footnote{The feature sets linguistic processes and punctuation correspond to the 'grammar' and punctuation feature set, respectively, in \cite{Rubin16}}

\mysubsection{\textit {Readability.}} We also extract features that indicate text understandability. These include content features such as the number of characters, complex words, long words, number of syllables, word types, and number of paragraphs, among others content features. We also calculate several readability metrics, including the  Flesch-Kincaid, Flesch Reading Ease, Gunning Fog, and the Automatic Readability Index (ARI).

\mysubsection{\textit{Syntax.}}  Finally, we extract a set of features derived  production rules based on context free grammars (CFG) trees using the Stanford Parser~\cite{Klein03}. The CFG derived features consist of all the lexicalized production rules (rules including child nodes) combined with their parent and grandparent node,  e.g., *NN\^{}NP$\rightarrow$commission (in this example NN --a noun-- is the grandparent node, NP --personal pronoun-- the parent node, and ``commissions" the child node. Features in this set are also encoded as tf-idf values.

\section{Computational Models for Fake News Detection}

We conduct several experiments with different (combinations of) feature sets. We use a linear SVM classifier and five-fold cross-validation, with accuracy, precision, recall, and F1 measures averaged over the five iterations. 

The machine learning classification was conducted with R \cite{R} and the caret \cite{caret} and e1071 packages \cite{e1071}.

Tables \ref{tab:results_amt} and \ref{tab:results_celeb} show the results obtained for the different feature sets.  
As seen in the tables, most of the classifiers obtain performances well above the random baseline of 0.50. The best performing classifier for the FakeNewsAMT dataset is derived from the \textit{Readability} features, followed by the combination of all linguistic feature sets. 
For the Celebrity dataset, the most accurate model is built using the \textit{Punctuation} features, followed by the \textit{Ngrams}, \textit{Complete LIWC}, and \textit{Syntax} features.

\begin{table*}[ht!]
\small
\centering
\begin{tabular}{l|l|ccc|ccc} 
                        \hline
                      \multirow{ 2}{*}{Features (number of features)} &   \multirow{ 2}{*}{Acc.}       & \multicolumn{3}{c|}{\sc Legitimate} & \multicolumn{3}{c}{\sc Fake}      \\ \cline{3-8}
                        &  & P      & R      & F1     & P       & R      & F1    \\ \hline
Punctuation (11)             & 0.70     & 0.67   & 0.77   & 0.72   & 0.73    & 0.63   & 0.68  \\
LIWC - Summary (7)  & 0.65     & 0.66   & 0.61   & 0.63    & 0.64    & 0.68    & 0.66     \\
LIWC - Linguistic processes (21)   & 0.64     & 0.64   & 0.63   & 0.63   & 0.63    & 0.64    & 0.63   \\
LIWC - Psychological processes (40)    & 0.58      & 0.58    & 0.58   & 0.58   & 0.58     & 0.57   & 0.57 \\
Complete LIWC (79)           & 0.67     & 0.68   & 0.66   & 0.67    & 0.67     & 0.68    & 0.67 \\ 
Readability  (26)     & 0.50     & 0.50   & 0.48   & 0.49    & 0.50     & 0.51    & 0.50    \\
Ngrams (1378)      & 0.67     & 0.67   & 0.66   & 0.66    & 0.66     & 0.68   & 0.67  \\
Syntax (1268)                    & 0.67     & 0.67   & 0.68   & 0.67    & 0.68     & 0.66    & 0.67       \\ \hline
All Features (2751)    & 0.73     & 0.73   & 0.72   & 0.72    & 0.73     & 0.74    & 0.73 \\\hline
\end{tabular}
\caption{Classification results for the Celebrity news data set.}
\label{tab:results_celeb}
\end{table*}

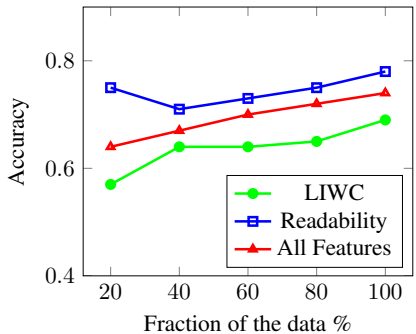
\begin{figure}[t]
\centering  
\begin{tikzpicture}[scale=\textwidth/20cm]

\begin{axis}[xlabel= Fraction of the data \% , ylabel=Accuracy, legend pos=south east,  ymin=0.4, ymax=.9, ystep=1]

\addplot [solid, every mark/.append style={solid, fill=green}, mark=*, color=green, very thick] table {
20	0.57
40	0.64
60	0.64
80	0.65
100	0.69
};
\addplot [solid, every mark/.append style={solid, fill=blue}, mark=square, color=blue, very thick]     
table {
20	0.75
40	0.71
60	0.73
80	0.75
100	0.78
};
\addplot [solid, every mark/.append style={solid, fill=red}, mark=triangle, color=red, very thick]     table {
20	0.64
40	0.67
60	0.7
80	0.72
100	0.74
};
\legend{LIWC,Readability,All Features}
\end{axis}
\end{tikzpicture}
\caption{Learning curves on the FakeNewsAMT dataset using three feature sets}
    \label{fig:learningCFakeAMT}
\end{figure}

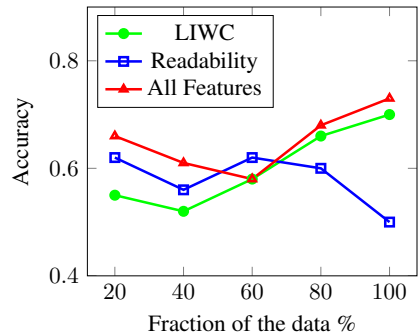
\begin{figure}[t]
    \centering
\begin{tikzpicture}[scale=\textwidth/20cm]

\begin{axis}[xlabel= Fraction of the data \% , ylabel=Accuracy, legend pos=north west,ymin=0.4, ymax=.9,ystep=1]

\addplot [solid, every mark/.append style={solid, fill=green}, mark=*, color=green, very thick] table {
20	0.55
40	0.52
60	0.58
80	0.66
100	0.7
};
\addplot [solid, every mark/.append style={solid, fill=blue}, mark=square, color=blue, very thick]     
table {
20	0.62
40	0.56
60	0.62
80	0.60
100	0.50

 };
\addplot [solid, every mark/.append style={solid, fill=red}, mark=triangle, color=red, very thick]     table {
20	0.66
40	0.61
60	0.58
80	0.68
100	0.73
};
\legend{LIWC,Readability, All Features}
\end{axis}
\end{tikzpicture}
\caption{Learning curves on the Celebrity dataset using three feature sets}
    \label{fig:learningCCeleb}
\end{figure}

\mysubsection{Learning Curves.} Next, we investigate whether larger amounts of training data can improve the identification of fake content. We plot the learning curves of the bests sets of features using incremental amounts of data as shown in Figures \ref{fig:learningCFakeAMT} and \ref{fig:learningCCeleb}. Except for the decrease obtained with the \textit{Readability} features on the Celebrity dataset, the learning trend for all the other feature sets on both datasets show steady improvement, thus suggesting that larger quantities of training data could improve the classification performance.

\begin{table*}[ht]
\small
\centering
    \begin{tabular}{l|l|l|l|cc}
    \hline
    Training  & Testing  & Feature set   & Acc. & F1$_{Legitimate}$  & F1$_{Fake}$ \\
    \hline
    \multirow{3}{*}{Celebrity}     & \multirow{3}{*}{FakeNewsAMT}  & Complete LIWC & 0.60 & 0.62           & 0.57    \\
         &   & Readability   & 0.61 & 0.60           & 0.67    \\
         &   & All Features   &  0.56 & 0.63           & 0.47     \\ \hline
    \multirow{3}{*}{FakeNewsAMT}   & \multirow{3}{*}{Celebrity}    & Complete LIWC & 0.61 & 0.62           & 0.57    \\
      &     & Readability   & 0.51 & 0.67           & 0.06    \\
      &     & All Features   &  0.51 &  0.67 & 0.08     \\
    \hline
    \end{tabular}
    \caption {Cross-domain analysis for best performing feature sets}
    \label{tab:crossdomain1}
\end{table*}

\mysubsection{Cross-domain Analyses.} We also explore the applicability of our methods across domains, using the two best feature sets ({\it Readability} and \textit{Complete LIWC}), as well as the classifier relying on all the features ({\it All Features}).  
Table \ref{tab:crossdomain1} shows the results obtained in cross-domain experiments between the FakeNewsAMT dataset and the Celebrity dataset. Perhaps not surprisingly, there is a significant loss in accuracy as compared to the within-domain results shown in Tables \ref{tab:results_amt} and \ref{tab:results_celeb}.

\begin{table*}[ht!]
\centering
\small
    \begin{tabular}{l|ccc|ccc|ccc}
    \hline
    \multirow{2}{*}{Domain} & \multicolumn{3}{c}{Readability}   & \multicolumn{3}{c}{Complete LIWC} & \multicolumn{3}{c}{All features} \\ \cline{2-10}
    & Acc. & F1$_{Legitimate}$ & F1$_{Fake}$ & Acc.& F1$_{Legitimate}$ & F1$_{Fake}$  & Acc.& F1$_{Legitimate}$ & F1$_{Fake}$ \\
    \hline
    Technology     & 0.90  &   0.90    &    0.90   &   0.62    &   0.57    &   0.64 & 0.80 & 0.78 & 0.81 \\
    Education      & 0.84   &  0.86    &   0.81    &  0.68   &    0.66    & 0.69 & 0.84    & 0.84    & 0.83\\
    Business       & 0.53   &   0.14    &   0.67    & 0.76   &  0.75    &   0.77 & 0.85    &   0.84    &   0.86\\
    Sports         & 0.51   &   0.26    &   0.64    & 0.73   &  0.74    &   0.70  &    0.81    &    0.81    &   0.81\\
    Politics       & 0.91   &  0.92    &   0.90    &  0.73   &    0.73    &   0.73  &  0.75    &   0.75    &   0.75 \\
    Entertainment  & 0.61   &   0.51    & 0.68  & 0.70   & 0.71 &   0.69  & 0.75    &   0.74 &  0.76 \\
    \hline
    \end{tabular}
\caption{Cross-domain classification accuracy for the complete LIWC and readability feature sets}
 \label{tab:crossdomain2}    
\end{table*}

The metrics suggest that the generalization from the crowdsourced data to the celebrity news is biased towards the truth (i.e., the classifier almost exclusively predicted the 'true' class). Possible explanations for the drop in performance might be (1) that the linguistic properties of deception in one domain are structurally different from those of deception in a second domain, and (2) that the feature sets applied for the cross-domain evaluation, in particular the readability feature set (accuracy = 0.50), were not performing well in the respective domain in the first place. To test this idea, we also applied cross-domain evaluation where we trained the classifier of domain A on the with the best feature set of domain B and tested in on domain B (here: \textit{Complete LIWC} and \textit{Readability} for the Celebrity and FakeNewsAMT data, respectively). The readability feature set classifier of the Celebrity data yielded an accuracy of 0.61 on the FakeNewsAMT data (compared to the original 0.78), and, vice versa, the \textit{Complete LIWC} classifier resulted in an accuracy of 0.61 (compared to 0.70). These findings indicate that different linguistic properties underlying different kinds of deception are more likely to explain cross-domain performance decreases than poorly performing feature sets.

We also assess the cross-domain classification performance for the six news domains in the FakeNewsAMT dataset. We do this by training on five of the six domains in the datset, and testing the remaining one. Table \ref{tab:crossdomain2} shows the results obtained in these experiments. The politics, education, and technology domains appear to be rather robust against classifiers trained on other domains. The technology and politics domains, moreover, are classified both with a high accuracy of 0.91 with the {\it Readability} feature set, which may suggest that fake and legitimate news in each of these three domains might be structurally similar to the fake and legitimate content in the other five domains. By contrast, domains such as sports, business and entertainment are less generalizable and might therefore be more domain-dependent. Although further research is needed to consolidate these findings, a possible explanation could be the rather unique content and style of these domains 

\section{Human Performance}

To identify a human baseline for the fake news detection task, we conducted a study to evaluate the human ability to spot fake news on the two developed datasets. We created an annotation interface that shows an annotator either a fake or a legitimate news article, and asks them to judge its credibility. We asked annotators to select a label of either ``Fake'' or ``Legitimate'' according to their own perceptions. We also asked them to indicate whether or not they have read or heard about the presented news in the past; overall, the annotators read less than 5\% of the news before, which we considered to be a negligible fraction. 

\begin{table}[t!]
\centering
\small
\begin{tabular}{l|c|c}\hline 
	&	Agreement	&	Kappa \\ \hline 
FakeNewsAMT & 70\% & 0.38\\
Celebrity & 73\% & 0.45\\ \hline
\end{tabular}
\caption{Agreement among two human annotators on the FakeNewsAMT and the Celebrity datasets.}
 \label{tab:agreementHA}
\end{table}

\begin{table}[t!]
\centering
\small
\begin{tabular}{l|cc}
 \hline
   & FakeNewsAMT  & Celebrity \\ \hline 
A1	&	0.71	&	0.80	\\
A2	&	0.70	&	0.77	\\
\hline
Sys	&	0.74	&	0.73	\\ \hline
\end{tabular}

\caption{Performance of two annotators (A1, A2) and the developed automatic system (Sys) on the fake news datasets}
 \label{tab:humanvsSys}
\end{table}

Two annotators labeled the news in each dataset. In both cases, the news articles were presented in a random order to avoid annotation bias. Annotators evaluated 480 and 200 news for the FakeNewsAMT and Celebrity datasets respectively. Annotators were not offered a monetary reward and we consider their judgments to be honest as they participated voluntarily in this experiment. Table \ref{tab:agreementHA} shows the observed agreement and Kappa statistics for each dataset.  Resulting Kappa values show moderate agreement values with slightly lower Kappa for the FakeNewAMT dataset. The results suggest that humans are better at identifying fake news in the celebrity domain than fake news in other domains.

In addition, we evaluate the performance of the automatic fake news classifiers against the human capability to spot fake news. Thus, we compare the accuracy of our system to that of human annotators. Table \ref{tab:humanvsSys} summarizes the accuracies obtained by the human annotators and our system on the two fake news datasets. Results confirm that humans are better at detecting fake content in the Celebrity domain. Notably, our system outperforms humans while detecting fake news in more serious and diverse news sources.

\section{Further Insights}

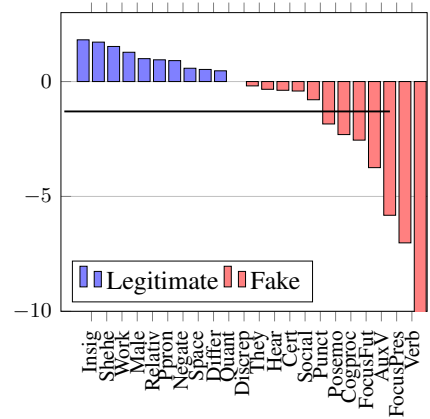
\begin{figure}[t!]
\centering
\begin{tikzpicture}[scale=\textwidth/18cm]
    \begin{axis}[x tick label style={/pgf/number format/1000 sep=},
        legend style={at={(0.5,-0.15)},anchor=north,legend columns=-1},
        ybar,
        legend pos=south west,
        xticklabels={ Insig, Shehe, Work, Male, Relativ, Ppron, Negate, Space, Differ, Quant, Discrep, They, Hear, Cert, Social,Punct, Posemo, Cogproc, FocusFut, AuxV, FocusPres, Verb},
        xtick={1,...,22},
        bar width=5pt,
        ymajorgrids=true,
        ymin=-10, ymax=3,ystep=2,
        tick label style={font=\small},
        xticklabel style={rotate=90},
after end axis/.append code={
   \draw[thick] (current axis.left of origin) -- (current axis.right of origin);
 },
    ]
        \addplot [draw=black,
            fill=blue!50,
        ] 
            coordinates { 
            (1,1.816)
(2,1.716)
(3,1.527)
(4,1.275)
(5,0.997)
(6,0.946)
(7,0.911)
(8,0.579)
(9,0.527)
(10,0.467)
};
        \addplot [draw=black,
            fill=red!50,
        ] 
            coordinates { 
            (11,-0.189)
(12,-0.337)
(13,-0.380)
(14,-0.412)
(15,-0.792)
(16,-1.846)
(17,-2.308)
(18,-2.550)
(19,-3.747)
(20,-5.820)
(21,-7.021)
(22,-11.723)

};
        \legend{Legitimate,Fake}
        
    \end{axis}
\end{tikzpicture}
\caption{\label{fig:liwcFake1} Language differences in fake and legitimate content in the FakeNewsAMT dataset}
\end{figure}

\begin{figure}[t!]
\centering
\begin{tikzpicture}[scale=\textwidth/18cm]
    \begin{axis}[x tick label style={/pgf/number format/1000 sep=},
        legend style={at={(0.5,-0.15)},anchor=north,legend columns=-1},
        ybar,
        legend pos=south west,
        xticklabels={Cogproc, Verb, FocusPres, Auxverb, Shehe, Negemo, Female, Differ, Discrep, Cause, Family, Netspeak, Hear, Compare, See, Posemo, Percept, I, Time, Prep, Leisure, Relativ},
        xtick={1,...,22},
        bar width=5pt,
        ymajorgrids=true,
        ymin=-2, ymax=3,ystep=2,
        tick label style={font=\small},
        xticklabel style={rotate=90},
after end axis/.append code={
   \draw[thick] (current axis.left of origin) -- (current axis.right of origin);
 },
    ]
        \addplot [draw=black,
            fill=blue!50,
        ] 
            coordinates { 
           ( 1,1.5232797198)
( 2,1.2068724243)
( 3,1.1460590096)
( 4,0.9562507367)
( 5,0.6757912785)
( 6,0.6106240641)
( 7,0.5029625166)
( 8,0.4066363264)
( 9,0.3466938803)
( 10,0.3233205915)
( 11,0.2812801181)
( 12,-0.0007083277)
};
        \addplot [draw=black,
            fill=red!50,
        ] 
            coordinates { 
            ( 12,-0.0007083277)
( 13,-0.2350597536)
( 14,-0.2501665912)
( 15,-0.3388433858)
( 16,-0.3952407019)
( 17,-0.5390173467)
( 18,-0.5478069747)
( 19,-0.5484790833)
( 20,-0.6142223838)
( 21,-0.6818671181)
( 22,-0.9185429564)

};
        \legend{Legitimate,Fake}
        
    \end{axis}
\end{tikzpicture}
\caption{\label{fig:liwcCelebrity} Language differences in fake and legitimate content in the Celebrity dataset}
\end{figure}
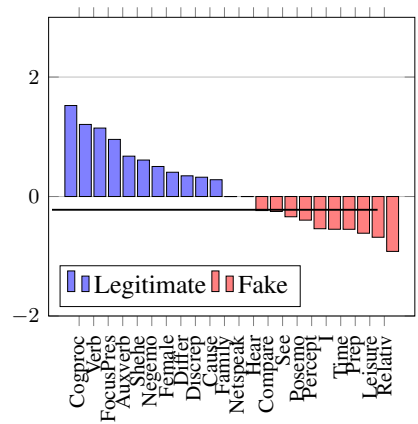

Our experiments suggest important differences in fake news content as compared to legitimate news content. Particularly, we observe that classifiers relying on the semantic information encoded in the LIWC lexicon show consistently good performance across domains. 
To gain further insights into the semantic classes that are associated with fake and legitimate content, we evaluate which classes show significant differences between the two groups of news. To compare both types of content, we subtract the average percentage of words in each LIWC category in the fake news from its corresponding values in the legitimate news set. Therefore, a positive result indicates an association between a LIWC class and legitimate content, and a negative result indicates an association between a LIWC class and fake content. Results for the FakeNewsAMT and Celebrity datasets  are shown in Figures \ref{fig:liwcFake1} and \ref{fig:liwcCelebrity} respectively. All the differences shown in the graphs are statically significant (one tailed t-test, p $<$ 0.5).

Figure \ref{fig:liwcFake1} indicates that the language used to report legitimate content in the FakeNewsAMT dataset, often includes words associated with cognitive processes such as insight and differentiation. In addition, legitimate content includes more function words such as he, she, and negations, and expresses relativity. On the other hand, language used when reporting fake content uses more social and positive words, expresses more certainty and focuses on present and future actions. Moreover, the authors of fake news use more adverbs, verbs, and punctuation characters than the authors of legitimate news. Likewise, results in Figure \ref{fig:liwcCelebrity} show noticeable differences among legitimate and fake content on the celebrity domain. Specifically, fake content in tabloid and entertainment magazines seem to use more perceptual words, e.g., hear, see, feeling, and positive emotions categories. In addition, fake content in this domain has a predominant use of the ``I'' pronoun and prepositions. In contrast, legitimate content uses words that indicate cognitive processes such as insight, cause, discrepancy, and tentative language. 

\section{Conclusions}

In this paper, we addressed the task of automatic identification of fake news. We introduced  two new fake news datasets, one obtained through crowdsourcing and covering six news domains, and another one obtained from the web covering celebrities.  We developed  classification models that rely on a combination of lexical, syntactic, and semantic information, as well features representing text readability properties. Our best performing models achieved accuracies that are comparable to human ability to spot fake content. 
\bibliographystyle{emnlp_natbib}

\bibliography{mainArxv.bib}
\end{document}